\documentclass[10pt,twocolumn,letterpaper]{article}

\usepackage[pagenumbers]{cvpr} % To force page numbers, e.g. for an arXiv version

\definecolor{cvprblue}{rgb}{0.21,0.49,0.74}
\usepackage[pagebackref,breaklinks,colorlinks,allcolors=cvprblue]{hyperref}
\usepackage{multirow}
\def\paperID{*****} % *** Enter the Paper ID here
\def\confName{CVPR}
\def\confYear{2026}

\title{All in One: Unifying Deepfake Detection, Tampering Localization, and Source Tracing with a Robust Landmark-Identity Watermark}

\usepackage{marvosym}
\author{
{Junjiang Wu}\textsuperscript{1}, Liejun Wang\textsuperscript{1,2 \Letter}, Zhiqing Guo\textsuperscript{1,2 \Letter}\\
\textsuperscript{1} School of Computer Science and Technology, Xinjiang University, Urumqi, China\\
\textsuperscript{2} Xinjiang Multimodal Intelligent Processing and Information Security \\ 
Engineering Technology Research Center, Urumqi, China\\
}

\usepackage{booktabs}
\usepackage{amsmath}
\usepackage{siunitx}
\usepackage{makecell}
\usepackage{pifont}

\begin{document}
\maketitle
{
    \renewcommand{\thefootnote}{}
    \footnotetext[0]{\Letter~Corresponding authors.}
}

\begin{abstract}
With the rapid advancement of deepfake technology, malicious face manipulations pose a significant threat to personal privacy and social security. However, existing proactive forensics methods typically treat deepfake detection, tampering localization, and source tracing as independent tasks, lacking a unified framework to address them jointly. To bridge this gap, we propose a unified proactive forensics framework that jointly addresses these three core tasks. Our core framework adopts an innovative 152-dimensional \textbf{l}andmark-\textbf{id}entity water\textbf{mark} termed \textbf{LIDMark}, which structurally interweaves facial landmarks with a unique source identifier. To robustly extract the LIDMark, we design a novel Factorized-Head Decoder (\textbf{FHD}). Its architecture factorizes the shared backbone features into two specialized heads (i.e., regression and classification), robustly reconstructing the embedded landmarks and identifier, respectively, even when subjected to severe distortion or tampering. This design realizes an ``all-in-one'' trifunctional forensic solution: the regression head underlies an ``intrinsic-extrinsic'' consistency check for detection and localization, while the classification head robustly decodes the source identifier for tracing. Extensive experiments show that the proposed LIDMark framework provides a unified, robust, and imperceptible solution for the detection, localization, and tracing of deepfake content. The code is available at \textcolor{blue}{https://github.com/vpsg-research/LIDMark}.
\end{abstract}

\begin{figure}[t!]
    \centering
    \includegraphics[width=\linewidth]{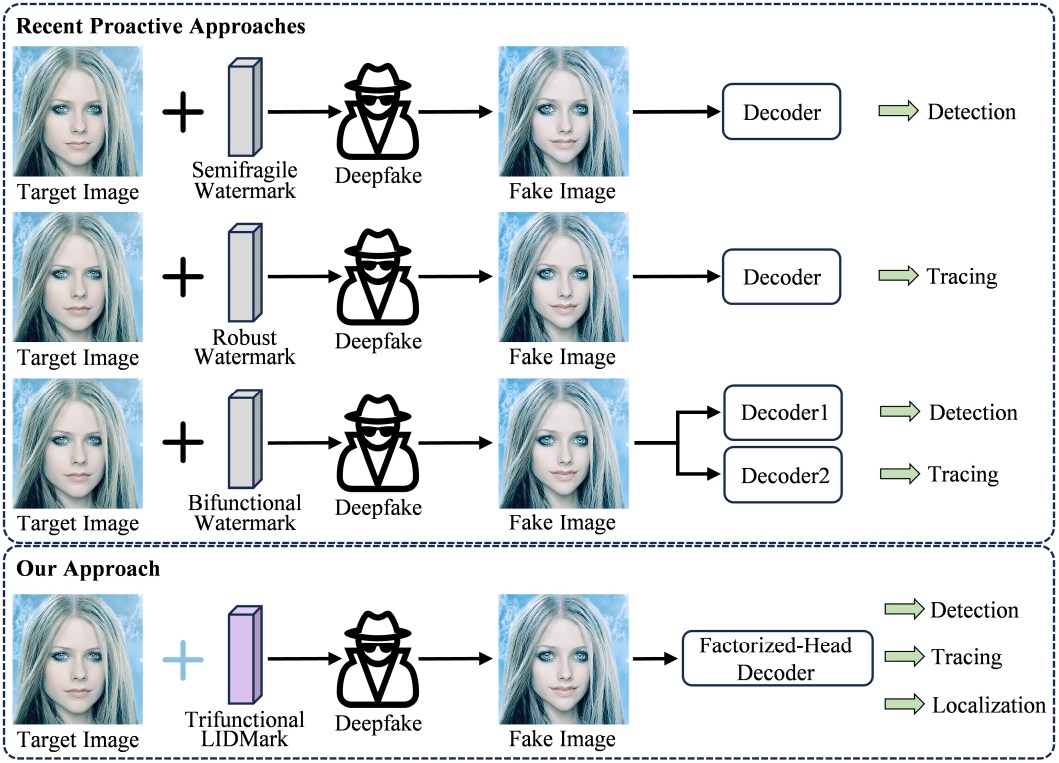}
    \caption{Comparison of proactive deepfake forensic paradigms. Conventional approaches shown at the top are limited to single tasks or require complex dual-decoder architectures for bifunctional forensics. The proposed ``all-in-one'' framework illustrated at the bottom employs the trifunctional LIDMark and a novel FHD for deepfake detection, source tracing, and tampering localization.}
    \label{fig:paradigm_comparison}
\end{figure}    
\section{Introduction}
\label{sec:intro}

The rapid development of generative models has significantly increased the sophistication and realism of deepfakes \cite{goodfellow2014generative, guo2023constructing, fernando2025face}. Although these technologies provide promising applications in movies, virtual reality and digital entertainment, their potential malicious applications pose unprecedented security risks and ethical dilemmas \cite{tolosana2020deepfakes, mirsky2021creation}. The high-fidelity visual content generated via deepfake techniques is extremely deceptive. When used maliciously, it becomes a powerful tool for fabricating false information, committing financial fraud and malicious impersonation, which poses a serious threat to public trust and social order. This erosion of ``seeing is believing'' makes it urgent to develop robust forensics technology to combat these escalating threats \cite{rossler2019faceforensics++, groh2022deepfake, ba2024exposing}. Conventional passive forensics methods seek to identify forgery artifacts from potentially manipulated content \cite{feng2025deepfake, gu2024deepfake, wang2024deepfake, guo2023ldfnet, guo2023exposing}. However, they are locked in a reactive arms race and consequently exhibit poor generalization against unseen manipulation techniques \cite{tolosana2020deepfakes}. These limitations have motivated the development of proactive methods, which preemptively embed imperceptible signals into target images to either disrupt the malicious generation process or enable downstream forensic tasks.

Current proactive approaches generally fall into two distinct paradigms: proactive defense and proactive forensics. Proactive defense employs adversarial perturbations \cite{szegedy2013intriguing, ruiz2020disrupting} to disrupt or suppress the deepfake generation process, but this approach has the risk of interfering with benign and legitimate generative applications. In contrast, proactive forensics technology embeds protection information through deep watermarking techniques \cite{uchida2017embedding, zhu2018hidden} to realize subsequent forensics tasks. The function of these watermarks is dictated by their designed robustness. Specifically, semi-fragile watermarks \cite{yang2021faceguard, neekhara2024facesigns} are designed to be sensitive to deepfake manipulation, enabling deepfake detection by verifying their integrity. Conversely, robust watermarks \cite{wang2021faketagger, wang2023robust, jia2025uncovering} are designed to be robust against deepfake manipulation, thereby surviving the deepfake pipeline and enabling the extraction of the embedded source identifier for source tracing. More recent bifunctional approaches \cite{wu2023sepmark, he2025kad, 11224900} attempt to combine these capabilities, but employ dual-decoder architectures to manage watermarks of differing robustness, thereby realizing the dual functions of detection and tracing. The difference between single-function and dual-function frameworks is shown at the top of Fig.~\ref{fig:paradigm_comparison}. However, the limitations of existing methods are twofold: technical and functional. Technically, they struggle to resolve the classic trade-off between robustness, imperceptibility, and high payload capacity \cite{qureshi2021detecting, bui2025trustmark, xu2025invismark}. High payload capacity, in particular, is critical for embedding diverse forensic information, a requirement that conventional low-bitrate identifiers cannot meet. Functionally, they suffer from a critical gap: the neglect of tampering localization. Current bifunctional methods, while focusing on detection (i.e., verifying authenticity) and tracing (i.e., identifying the source), fail to answer the crucial third question: Where is it fake? The ability to pinpoint which facial regions are manipulated is crucial, as it moves beyond a simple ``real or fake'' verdict and enables a granular assessment of the content's semantic integrity, i.e., understanding precisely which facial components were altered. This localization data is indispensable for advanced forensic analysis, such as inferring the malicious manipulation's intent or providing objective evidence to substantiate the detection verdict.

These identified limitations motivate our pursuit of a truly comprehensive, trifunctional proactive forensics solution. We argue that an ideal framework must move beyond paired tasks and concurrently address three fundamental forensic questions: (1) Deepfake Detection: Is the image authentic or forged? (2) Tampering Localization: If forged, which specific facial regions have been manipulated? (3) Source Tracing: Whether the image is real or tampered with, can the unique identifier of the image source be reliably recovered? While existing methods tackle these as single or paired problems (Fig.~\ref{fig:paradigm_comparison}, top), a unified framework that simultaneously achieves all three remains a significant and open challenge. To the best of our knowledge, no single method provides this ``all-in-one'' forensic solution.

Inspired by this insight, we propose a unified proactive forensics framework centered on our trifunctional 152-D \textbf{l}andmark-\textbf{id}entity water\textbf{mark}, termed \textbf{LIDMark}, which is illustrated at the bottom of Fig.~\ref{fig:paradigm_comparison}. LIDMark is the first to structurally interweave two distinct payloads: (1) a tamper-sensitive 136-D facial landmark vector and (2) a robust 16-D source identifier. To robustly decode the LIDMark, we design a novel Factorized-Head Decoder (FHD). Different from the complex dual-decoder architecture in previous studies, FHD reconstructs heterogeneous landmarks and identity payloads synchronously by decomposing the shared backbone features into two dedicated heads for regression and classification. This unified design realizes an ``all-in-one'' trifunctional forensic solution: for detection and localization, we introduce a mechanism called the ``intrinsic-extrinsic'' consistency check \cite{wang2024lampmark}. This check compares the intrinsic landmarks (i.e., the original facial geometry reconstructed by the FHD's regression head) against the extrinsic landmarks (i.e., the facial geometry re-detected from the potentially manipulated image). Globally, a high Average Euclidean Distance (AED) indicates forgery, while locally, the semantic structure of our landmarks enables a region-based inconsistency analysis to pinpoint tampered regions. Concurrently, for source tracing, the FHD's classification head robustly decodes the embedded source identifier. The main contributions of this paper are as follows:

\begin{itemize}
    \item We propose a novel 152-D landmark-identity watermark, \textbf{LIDMark}, the first to structurally interweave a 136-D tamper-sensitive facial landmark vector with a 16-D robust source identifier into a single and unified payload.
    
    \item We design a novel Factorized-Head Decoder, termed FHD, to factorize shared backbone features into specialized regression and classification heads, concurrently and robustly reconstructing the LIDMark's heterogeneous facial landmarks and source identifier.
    
    \item We present a unified, trifunctional proactive forensics solution, the LIDMark framework, which is the first to realize deepfake detection, tampering localization and source tracing jointly. This ``all-in-one'' solution is realized by our LIDMark, FHD, and a mechanism called the ``intrinsic-extrinsic'' consistency check.
\end{itemize}
\section{Related Work}
\label{sec:formatting}

\subsection{Proactive Defense}

Proactive defense injects imperceptible perturbations into a target image, causing the generation process to fail by forcing the model to produce a non-malicious or visually corrupted output. For example, Ruiz et al. \cite{ruiz2020disrupting} proposed an adversarial attack method that utilizes imperceptible perturbations to interfere with the generative translation process. Subsequent work has focused on improving the robustness of these perturbations, such as using perceptual-aware noise in the Lab color space \cite{ijcai2022p107} or developing GAN-based methods to resist online social network compression \cite{qu2024df}. However, the ``disrupt-first'' paradigm is limited in that it lacks key forensic functionalities. While recent nascent ``traceable'' approaches \cite{zhang2024dual, li2024dual} attempt to bridge this gap, they remain fundamentally fragile and offer insufficient capacity for meaningful forensic analysis. Furthermore, by retaining disruption as their primary goal, all the aforementioned methods risk interfering with benign applications.

\begin{figure}[t!]
    \centering
    \includegraphics[width=\linewidth]{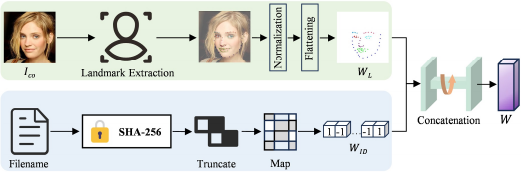}
    \caption{The 152-D LIDMark $W$ construction. The composite watermark concatenates two primary streams: (1) $W_L$, a 136-D vector of normalized 2-D facial landmarks, and (2) $W_{ID}$, a 16-D bipolar identifier derived via a SHA-256 hash of the filename.}
    \label{fig:composition}
\end{figure}
\subsection{Proactive Forensics}

Proactive forensics employs deep watermarking to embed forensic information, whose designed fragility determines its function. This leads to three lines of work. First, semi-fragile watermarking leverages fragility for deepfake detection. FaceSigns \cite{neekhara2024facesigns} employs a deep learning-based semi-fragile framework where watermark breakage signifies a deepfake manipulation. Moreover, Asnani et al. \cite{asnani2022proactive} propose a proactive scheme that embeds learnable templates into the target image. Detection is then achieved by measuring the differential recoverability of these templates from the real image versus the manipulated version. Second, robust watermarking targets source tracing by ensuring the watermark survives the deepfake pipeline. FakeTagger \cite{wang2021faketagger} utilizes an encoder-decoder architecture with channel coding to ensure robust recovery of an embedded message after GAN-based transformations. More recently, DiffMark \cite{SUN2026103801} leverages a diffusion model as the embedder, guiding the denoising process with the image and watermark as conditions to produce a watermarked image exhibiting high robustness. Finally, most advanced methods increasingly seek dual-functionality. A mainstream strategy is the dual-decoder architecture. SepMark \cite{wu2023sepmark} introduces separable watermarking, which extracts a watermark using two decoders tuned to different robustness levels: a semi-fragile ``Detector'' for detection and a robust ``Tracer'' for tracing. KAD-Net \cite{he2025kad} employs a similar decoupled dual-branch framework, with its differential-aware module providing sensitivity to manipulation and its tokenized Kolmogorov-Arnold network ensuring robustness for tracing. WaveGuard \cite{11224900} also employs a separable decoder architecture that leverages different DT-CWT sub-band combinations to achieve sensitive detection and robust tracing, respectively. Despite these advances, the field still lacks a unified framework for simultaneous detection, localization, and tracing from a single forensic payload. This gap motivates the development of a more integrated and trifunctional forensic solution.

\begin{figure*}[t!]
    \centering
    \includegraphics[width=\linewidth]{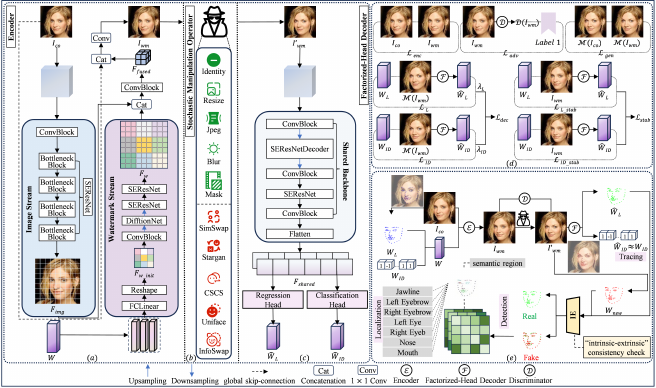}
    \caption{Overview of the LIDMark framework. The trifunctional forensic framework features an encoder $E$, a stochastic manipulation operator $\mathcal{M}$, a factorized-head decoder FHD, and a discriminator $D$. (a) The encoder embeds the LIDMark $W$ into $I_{co}$ via a two-stream fusion network, yielding the watermarked image $I_{wm}$. (b) $\mathcal{M}$ simulates diverse common distortions and deepfake manipulations on $I_{wm}$, producing the manipulated image $I'_{wm}$. (c) The FHD recovers $\hat{W}_L$ and $\hat{W}_{ID}$ from $I'_{wm}$ via the shared backbone and factorized heads. (d) The multi-task loss functions guide the training process. (e) The ``intrinsic-extrinsic'' consistency check employs recovered landmarks $\hat{W}_L$ and re-detected landmarks $W_{new}$ for fine-grained tamper detection and localization, while identifiers $\hat{W}_{ID}$ are extracted for source tracing.}
    \label{fig:architecture}
\end{figure*}
\section{Methods}

\subsection{Overview of LIDMark}
\label{sec:LIDMark_Composition}
We propose LIDMark $W$, a purely robust $152$-D composite watermark as a unified payload for three tasks. As illustrated in Fig.~\ref{fig:composition}, $W$ concatenates two heterogeneous information streams. The first component, landmark vector ($W_L \in [0, 1]^{136}$), encodes fine-grained facial geometry, serving as the primary forensic signal for detection and localization. We employ the face-alignment library \cite{bulat2017far} to extract 68 facial landmarks, $\mathcal{L} = \{(x_i, y_i)\}_{i=1}^{68}$. To ensure scale invariance, we normalize their coordinates by the image width $w$ and height $h$, yielding $(\tilde{x}_i, \tilde{y}_i) = (\frac{x_i}{w}, \frac{y_i}{h})$. These normalized coordinates are flattened to form the 136-D vector $W_L$. The second component, identifier vector ($W_{ID} \in \{-1, 1\}^{16}$), is a bipolar vector for source tracing. To guarantee uniqueness and high collision-resistance, we employ a deterministic hash generation scheme. Specifically, the image filename serves as input to the SHA-256 algorithm. The resulting digest is truncated and mapped to the 16-D bipolar vector $W_{ID}$. The potential for future expansion of the identifier payload and a scalability assessment are detailed in Appendix A. This composite design, $W = [W_L ; W_{ID}]$, thus interweaves a tamper-sensitive geometric signal $W_L$ with a robust source identifier $W_{ID}$.

\subsection{Network Architecture}
\label{sec:Network_Architecture}
To imperceptibly embed and robustly recover LIDMark, we propose the end-to-end LIDMark framework. As shown in Fig.~\ref{fig:architecture}, the framework comprises an encoder, a novel factorized-head decoder, and a discriminator, jointly trained against a stochastic manipulation operator.

\textbf{Encoder Architecture.}
\label{sec:Encoder_Architecture}
The encoder $E$ is a two-stream fusion network, designed to embed $W$ into the cover image $I_{co}$ while preserving visual fidelity via a global skip-connection.
\textbf{Image Stream.} The input $I_{co}$ is fed into a backbone, composed of a ConvBlock and a SEResNet module, to extract semantic features $F_{img}$.
\textbf{Watermark Stream.} Concurrently, $W$ is mapped by a fully-connected (FC) layer to a 256-dimensional vector and reshaped into a $1 \times 16 \times 16$ map $F_{w\_init}$. This map is processed by a backbone comprising a ConvBlock, a DiffusionNet for spatial upsampling, and two sequential SEResNet modules. This process refines $F_{w\_init}$ and expands it into the watermark feature map $F_w$, matching the spatial resolution of $F_{img}$.
\textbf{Fusion and Reconstruction.} $F_{img}$ and $F_w$ are concatenated channel-wise and fused by a ConvBlock. This fused map is concatenated with $I_{co}$ in a global skip-connection. A final $1 \times 1$ convolution regresses this aggregated map to produce the watermarked image $I_{wm} = E(I_{co}, W)$.

\begin{figure*}[t]
    \centering
    \includegraphics[width=\linewidth]{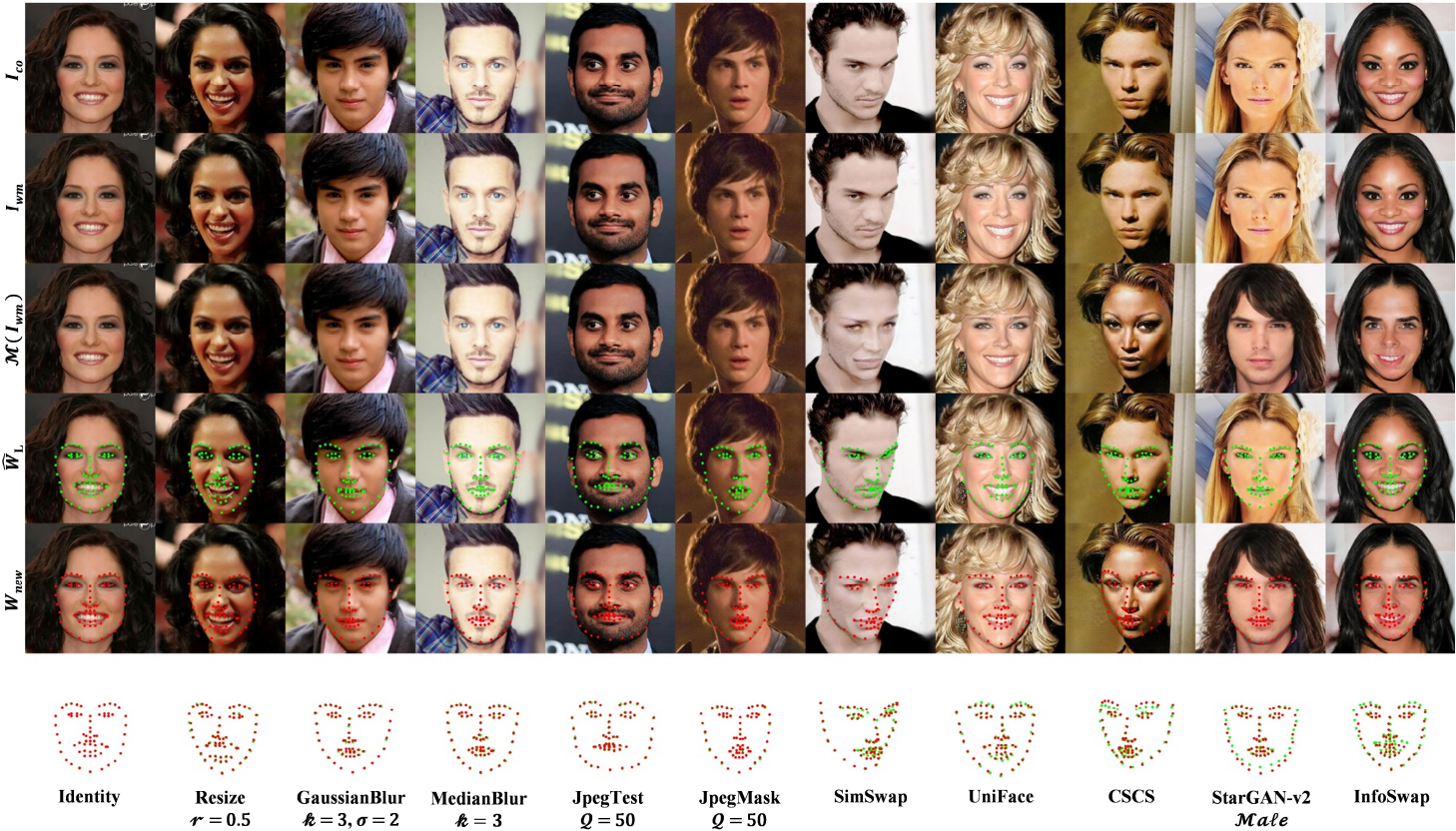}
    \caption{Visual assessment of LIDMark robustness and imperceptibility. Comparing rows 1 and 2 shows the watermarked image $I_{wm}$ is indistinguishable from the cover image $I_{co}$. Row 3 displays the manipulation results $\mathcal{M}(I_{wm})$. The ``intrinsic-extrinsic'' consistency check compares the green dots in row 4, representing FHD-recovered intrinsic landmarks $\hat{W}_L$, against the red dots in row 5, representing re-detected extrinsic landmarks $W_{new}$. Row 6 merges these two landmark sets into a combined image to visualize their spatial differences.}
    \label{fig:visualization}
\end{figure*}

\textbf{Stochastic Manipulation Operator.}
The image $I_{wm}$ is processed by the operator $\mathcal{M}$, yielding the stochastically manipulated image $I'_{wm} = \mathcal{M}(I_{wm})$. Instead of a fixed layer, $\mathcal{M}$ is a probabilistic operator simulating diverse attacks, prompting the FHD to generalize beyond any single distortion. In the pre-training stage, $\mathcal{M}_{c}$ samples from a pool of common distortions. This pool includes \textit{Identity}, \textit{Resize}, \textit{GausBlur}, \textit{MedBlur}, \textit{JpegTest}, and \textit{JpegMask}. The specific parameters of the above distortions are provided in Fig.~\ref{fig:visualization}. To evaluate robustness against deepfakes, we consider a comprehensive suite of five diverse models: SimSwap \cite{chen2020simswap}, UniFace \cite{xu2022designing}, CSCS \cite{huang2024identity}, StarGAN-v2 \cite{choi2020stargan}, and InfoSwap \cite{gao2021information}. During the fine-tuning stage, $\mathcal{M}_{d}$ samples exclusively from the first four, strictly reserving InfoSwap \cite{gao2021information} for the testing phase to evaluate generalization against unseen attacks. Simulating these attacks requires pairing a target image that provides pose and structure with a source image that supplies a new facial identity. We employ an efficient intra-batch rolling strategy, using another image from the same batch as the identity source. For both stages, we adopt a single-task-per-batch strategy \cite{zhang2021towards}: one manipulation is randomly selected and applied uniformly to all samples within a mini-batch. This approach forces the FHD to learn features specialized against the distinct artifacts of each attack, preventing convergence to a weak ``average-case'' solution ineffective against specific attacks.

\textbf{Factorized-Head Decoder.}
The FHD is a multi-task learning network designed to simultaneously recover the two heterogeneous watermark components, $W_L$ and $W_{ID}$, from the manipulated image $I'_{wm}$. Its architecture consists of a shared backbone and two factorized heads.
\textbf{Shared Backbone.} A convolutional backbone, comprising multiple ConvBlocks, a SEResNetDecoder, and a SEResNet module, extracts a shared latent representation $F_{shared}$ from $I'_{wm}$.
\textbf{Factorized Heads.} $F_{shared}$ is flattened and fed into two parallel FC heads to produce the network output $\text{FHD}(I'_{wm}) = (\hat{W}_L, \hat{W}_{ID})$. Specifically, these heads comprise: (i) a landmark regression head that outputs the 136-D continuous vector $\hat{W}_L \in \mathbb{R}^{136}$, and (ii) an identifier classification head that outputs the 16-D identifier logits $\hat{W}_{ID} \in \mathbb{R}^{16}$. This factorized design is crucial, as it decouples the disparate sub-tasks of landmark regression and identifier classification, allowing for their joint optimization while respecting the distinct mathematical nature of each.

\textbf{Discriminator Architecture.}
The discriminator $D$ is a deep convolutional classifier driving the adversarial objective. It processes either $I_{co}$ or $I_{wm}$ through a stack of ConvBlocks. A global average pooling layer then aggregates the spatial features into a vector, which is subsequently projected by a final FC layer to a single scalar logit.

\subsection{Multi-Task Loss Function}
\label{sec:Multi-Task_Loss Function}
The framework is optimized via a multi-task objective, as illustrated in Fig.~\ref{fig:architecture}(d). Throughout this section, $\lambda_{(\cdot)}$ denotes the balancing hyperparameter for its respective loss term.

\textbf{Encoder Loss ($\mathcal{L}_{enc}$).} To ensure imperceptibility, we minimize $\mathcal{L}_{enc}$:
%-------------------------
\begin{equation}
    \mathcal{L}_{enc} = \mathbb{E}[\|I_{wm} - I_{co}\|_2^2]
    \label{eq:loss_enc}
\end{equation}
%-------------------------

\textbf{FHD Recoverability Loss ($\mathcal{L}_{dec}$).} We jointly optimize the FHD heads via a combined decoding loss $\mathcal{L}_{dec}$, formed by two sub-losses:
(i) Landmark Regression Loss ($\mathcal{L}_{L}$). To preserve geometric structure, we minimize the mean point-wise Euclidean distance:
%-------------------------
\begin{equation}
    \mathcal{L}_{L} = \frac{1}{B \cdot N_{k}} \sum_{b=1}^{B} \sum_{i=1}^{N_{k}} \| \hat{p}_{b,i} - p_{b,i} \|_2
    \label{eq:loss_landmark}
\end{equation}
%-------------------------
where $B$ and $N_k=68$ denote the batch size and landmark count, while $p_{b,i}$ and $\hat{p}_{b,i}$ are the target and predicted 2D points.
(ii) Identifier Classification Loss ($\mathcal{L}_{ID}$). We apply the Binary Cross-Entropy (BCE) loss to the predicted logits $\hat{W}_{ID}$:
%-------------------------
\begin{equation}
    \mathcal{L}_{ID} = \text{BCE}(\sigma(\hat{W}_{ID}), \frac{W_{ID} + 1}{2})
    \label{eq:loss_id}
\end{equation}
%-------------------------
where $\sigma(\cdot)$ denotes the Sigmoid activation function.

The final decoding loss $\mathcal{L}_{dec}$ is a weighted sum of these sub-losses:
%-------------------------
\begin{equation}
    \mathcal{L}_{dec} = \lambda_{L} \mathcal{L}_{L} + \lambda_{ID} \mathcal{L}_{ID}
    \label{eq:loss_dec_total}
\end{equation}
%-------------------------

\textbf{Discriminator Loss ($\mathcal{L}_{D}$).} $\mathcal{L}_{D}$ trains $D$ to separate $I_{co}$ (label 1) and $I_{wm}$ (label 0). Meanwhile, $\mathcal{L}_{adv}$ trains $E$ to fool $D$ by driving $D(I_{wm})$ to 1:
%-------------------------
\begin{equation}
    \mathcal{L}_{D} = \mathbb{E}_{I_{co}}[\mathcal{L}_{BCE}(D(I_{co}), 1)] + \mathbb{E}_{I_{wm}}[\mathcal{L}_{BCE}(D(I_{wm}), 0)]
    \label{eq:loss_d}
\end{equation}
%-------------------------
%-------------------------
\begin{equation}
    \mathcal{L}_{adv} = \mathbb{E}_{I_{wm}}[\mathcal{L}_{BCE}(D(I_{wm}), 1)]
    \label{eq:loss_adv}
\end{equation}
%-------------------------

\begin{table}[t]
\centering
\caption{Objective imperceptibility comparison on CelebA-HQ \cite{karras2017progressive}. Results are shown across $128 \times 128$ and $256 \times 256$ resolutions. W.L. denotes the watermark length.}
\label{tab:imperceptibility}
\setlength{\tabcolsep}{4pt}
\setlength{\heavyrulewidth}{1.5pt}
\begin{tabular}{lcccc} 
\toprule
Model & Image Size & W.L. & PSNR $\uparrow$ & SSIM $\uparrow$ \\
\midrule
MBRS \cite{jia2021mbrs} & $128\times128$ & 30 & 35.19 & 0.90 \\
CIN \cite{ma2022towards} & $128\times128$ & 30 & 39.70 & 0.93 \\
SepMark \cite{wu2023sepmark} & $128\times128$ & 30 & 38.31 & 0.96 \\
EditGuard \cite{zhang2024editguard} & $128\times128$ & 30 & 37.17 & 0.95 \\
LampMark \cite{wang2024lampmark} & $128\times128$ & 30 & \underline{40.20} & \underline{0.97} \\
DiffMark \cite{SUN2026103801} & $128\times128$ & 30 & 40.12 & 0.96 \\
KAD-NET \cite{he2025kad} & $128\times128$ & 30 & 39.98 & 0.96 \\
\textbf{Ours} & $128\times128$ & \textbf{152} & \textbf{40.22} & \textbf{0.98} \\
\midrule
MBRS \cite{jia2021mbrs} & $256\times256$ & 128 & 36.34 & 0.89 \\
CIN \cite{ma2022towards} & $256\times256$ & 128 & 37.15 & 0.85 \\
SepMark \cite{wu2023sepmark} & $256\times256$ & 128 & 38.47 & 0.93 \\
EditGuard \cite{zhang2024editguard} & $256\times256$ & 128 & 36.66 & 0.89 \\
LampMark \cite{wang2024lampmark} & $256\times256$ & 128 & \underline{42.52} & 0.95 \\
DiffMark \cite{SUN2026103801} & $256\times256$ & 128 & 41.96 & \underline{0.98} \\
KAD-NET \cite{he2025kad} & $256\times256$ & 128 & 40.07 & 0.94 \\
\textbf{Ours} & $256\times256$ & \textbf{152} & \textbf{44.31} & \textbf{0.99} \\
\bottomrule
\end{tabular}
\end{table}

\textbf{Generative Consistency Loss ($\mathcal{L}_{gen}$)}. Exclusive to the fine-tuning stage, $\mathcal{L}_{gen}$ ensures the watermark is agnostic to $\mathcal{M}$:
%-------------------------
\begin{equation}
    \mathcal{L}_{gen} = \mathbb{E}[\|\mathcal{M}(I_{wm}) - \mathcal{M}(I_{co})\|_2^2]
    \label{eq:loss_gen}
\end{equation}
%-------------------------

\textbf{Overall Generator Loss ($\mathcal{L}_{G}$)}. The parameters of $E$ and FHD are jointly optimized via the loss $\mathcal{L}_{G}$. The formulation of this loss varies across our two-stage training strategy:

(i) \textbf{Pre-training Phase.} During pre-training on common manipulations, the total generator loss $\mathcal{L}_{G1}$ is defined as a weighted sum of three components:
%-------------------------
\begin{equation}
    \mathcal{L}_{G1} = \lambda_{enc} \mathcal{L}_{enc} + \mathcal{L}_{dec} + \lambda_{adv} \mathcal{L}_{adv}
    \label{eq:loss_total_1}
\end{equation}
%-------------------------

(ii) \textbf{Deepfake Fine-tuning Phase.} During fine-tuning on the selected deepfake operators $\mathcal{M}_{d}$, we introduce $\mathcal{L}_{gen}$ and a decoder stability loss $\mathcal{L}_{stab}$. To mitigate catastrophic forgetting, wherein the FHD fails to decode the non-attacked watermarked image $I_{wm}$, $\mathcal{L}_{stab}$ is computed on $I_{wm}$ as the unweighted sum of the recovery losses: $\mathcal{L}_{stab} = \mathcal{L}_{L\_stab} + \mathcal{L}_{ID\_stab}$. The total generator loss $\mathcal{L}_{G2}$ is formulated as:
\begin{equation}
    \mathcal{L}_{G2} = \lambda_{enc}\mathcal{L}_{enc} + \mathcal{L}_{dec} + \lambda_{adv}\mathcal{L}_{adv} + \lambda_{gen}\mathcal{L}_{gen} + \lambda_{stab}\mathcal{L}_{stab}
    \label{eq:loss_total_2}
\end{equation}
Since the regression task demands strict pixel-level accuracy while the classification task allows for a certain degree of deviation, we prioritize the former by setting $\lambda_{stab} = \lambda_{L}$.

\section{Experiments}
\label{sec:experiments}

\subsection{Implementation Details}
\label{sec:Implementation_Details}
\textbf{Datasets.}
Our primary experiments adopt the CelebA-HQ dataset \cite{karras2017progressive}, comprising 30,000 images. In line with leading forensic watermarking methods \cite{wu2023sepmark, wang2024lampmark, SUN2026103801, he2025kad}, this dataset serves as the main training and evaluation benchmark using official data splits. To assess generalization, we employ the LFW dataset \cite{huang2008labeled}, featuring 5,749 unique identities, and randomly select 2,000 images for testing. All images are resized to $128 \times 128$ and $256 \times 256$ to analyze performance across resolutions. The $152$-D LIDMark is pre-generated for each corresponding image, as detailed in Sec.~\ref{sec:LIDMark_Composition}.

\textbf{Training Strategy.}
The LIDMark framework is implemented in PyTorch and optimized on NVIDIA A40 GPUs using a two-stage strategy. (i) Pre-training: This phase is conducted on $\mathcal{M}_{c}$ for 100 epochs using the Adam optimizer. The batch size is set to 32, and the learning rate to $4.3 \times 10^{-4}$, minimizing $\mathcal{L}_{G1}$ in Eq.~\eqref{eq:loss_total_1} with loss weights $[\lambda_{L}, \lambda_{ID}] = [11.5, 14.7]$. (ii) Fine-tuning: Subsequent optimization on $\mathcal{M}_{d}$ is performed for another 100 epochs. Given the substantial GPU memory consumption of deepfake models, the batch size is reduced to 8. The learning rate is set to $4.0 \times 10^{-4}$ to optimize $\mathcal{L}_{G2}$ in Eq.~\eqref{eq:loss_total_2}, updating the balancing weights to $[\lambda_{L}, \lambda_{ID}] = [4.2, 1.0]$. A detailed ablation study of $\lambda_{L}$ and $\lambda_{ID}$ is provided in Sec.~\ref{sec:ablation_studies}.

\textbf{Baselines.}
To the best of our knowledge, LIDMark is the first framework to unify three distinct forensic tasks into a single model. Consequently, no direct trifunctional baselines exist. We benchmark our framework against representative single-function and dual-function methods, including MBRS \cite{jia2021mbrs}, CIN \cite{ma2022towards}, SepMark \cite{wu2023sepmark}, EditGuard \cite{zhang2024editguard}, LampMark \cite{wang2024lampmark}, DiffMark \cite{SUN2026103801}, and KAD-NET \cite{he2025kad}.

\begin{table*}[t]
\centering
\caption{Quantitative comparison of BER and AED on CelebA-HQ \cite{karras2017progressive} under common distortions $\mathcal{M}_{c}$. All results are reported at $128 \times 128$ and $256 \times 256$ resolutions. GausNoise and MedBlur denote Gaussian noise and median blur, respectively.}
\label{tab:common_robustness}
\resizebox{\textwidth}{!}{%
\begin{tabular}{l|cc|cc|cc|cc|cc|cc|cc|cccc} 
\Xhline{1.5pt}
\multirow{3}{*}{Distortion} & \multicolumn{2}{c|}{MBRS \cite{jia2021mbrs}} & \multicolumn{2}{c|}{CIN \cite{ma2022towards}} & \multicolumn{2}{c|}{SepMark \cite{wu2023sepmark}} & \multicolumn{2}{c|}{EditGuard \cite{zhang2024editguard}} & \multicolumn{2}{c|}{LampMark \cite{wang2024lampmark}} & \multicolumn{2}{c|}{DiffMark \cite{SUN2026103801}} & \multicolumn{2}{c|}{KAD-NET \cite{he2025kad}} & \multicolumn{4}{c}{\textbf{Ours}} \\ 
\cline{2-19}
 & \multicolumn{2}{c|}{BER} & \multicolumn{2}{c|}{BER} & \multicolumn{2}{c|}{BER} & \multicolumn{2}{c|}{BER} & \multicolumn{2}{c|}{BER} & \multicolumn{2}{c|}{BER} & \multicolumn{2}{c|}{BER} & \multicolumn{2}{c|}{BER} & \multicolumn{2}{c}{AED} \\ 
\cline{2-19}
 & 128 & 256 & 128 & 256 & 128 & 256 & 128 & 256 & 128 & 256 & 128 & 256 & 128 & 256 & 128 & 256 & 128 & 256 \\ 
\Xhline{0.6pt}
Identity & 0.00\% & 0.00\% & 0.00\% & 0.00\% & 0.00\% & 0.00\% & 0.09\% & 0.12\% & 0.00\% & 0.00\% & 0.00\% & 0.00\% & 0.00\% & 0.00\% & 0.00\% & 0.00\% & 1.06 & 2.96 \\
Resize & 10.72\% & 13.58\% & 24.97\% & 35.21\% & 23.81\% & 3.40\% & 0.60\% & 0.28\% & 14.00\% & 13.82\% & 0.02\% & 0.01\% & 0.00\% & 0.13\% & 0.00\% & 0.00\% & 1.19 & 3.00 \\
GausBlur & 4.57\% & 10.14\% & 6.89\% & 0.39\% & 0.41\% & 0.04\% & 1.53\% & 3.37\% & 1.82\% & 0.59\% & 0.00\% & 0.00\% & 0.00\% & 0.00\% & 0.00\% & 0.00\% & 1.05 & 3.03 \\
MedBlur & 1.11\% & 5.55\% & 0.81\% & 0.65\% & 0.20\% & 0.03\% & 1.62\% & 3.85\% & 1.72\% & 0.63\% & 0.00\% & 0.00\% & 0.00\% & 0.00\% & 0.00\% & 0.00\% & 1.06 & 3.01 \\
JpegTest & 0.00\% & 0.01\% & 5.25\% & 8.90\% & 1.22\% & 0.10\% & 1.42\% & 3.50\% & 0.69\% & 1.83\% & 0.85\% & 1.81\% & 1.81\% & 0.01\% & 1.62\% & 0.57\% & 1.57 & 3.39 \\
JpegMask & 0.85\% & 7.24\% & 3.69\% & 6.31\% & 0.39\% & 1.92\% & 1.21\% & 3.87\% & 0.26\% & 1.44\% & 1.88\% & 2.23\% & 1.92\% & 3.16\% & 0.00\% & 0.01\% & 1.08 & 2.98 \\
Average & 2.88\% & 6.09\% & 6.94\% & 8.58\% & 4.34\% & 0.92\% & 1.08\% & 2.50\% & 3.08\% & 3.05\% & \underline{0.46\%} & 0.68\% & 0.62\% & \underline{0.55\%} & \textbf{0.27}\% & \textbf{0.10}\% & \textbf{1.17} & \textbf{3.06} \\ 
\Xhline{1.5pt}
\end{tabular}
}
\end{table*}

\begin{table*}[t]
\centering
\caption{Quantitative comparison of BER and AED on CelebA-HQ \cite{karras2017progressive} under deepfake manipulations $\mathcal{M}_{d}$. All results are reported at $128 \times 128$ and $256 \times 256$ resolutions.}
\label{tab:deepfake_robustness}
\resizebox{\textwidth}{!}{%
\begin{tabular}{l|cc|cc|cc|cc|cc|cc|cc|cccc} 
\Xhline{1.5pt}
\multirow{3}{*}{Manipulation} & \multicolumn{2}{c|}{MBRS \cite{jia2021mbrs}} & \multicolumn{2}{c|}{CIN \cite{ma2022towards}} & \multicolumn{2}{c|}{SepMark \cite{wu2023sepmark}} & \multicolumn{2}{c|}{EditGuard \cite{zhang2024editguard}} & \multicolumn{2}{c|}{LampMark \cite{wang2024lampmark}} & \multicolumn{2}{c|}{DiffMark \cite{SUN2026103801}} & \multicolumn{2}{c|}{KAD-NET \cite{he2025kad}} & \multicolumn{4}{c}{\textbf{Ours}} \\ 
\cline{2-19}
 & \multicolumn{2}{c|}{BER} & \multicolumn{2}{c|}{BER} & \multicolumn{2}{c|}{BER} & \multicolumn{2}{c|}{BER} & \multicolumn{2}{c|}{BER} & \multicolumn{2}{c|}{BER} & \multicolumn{2}{c|}{BER} & \multicolumn{2}{c|}{BER} & \multicolumn{2}{c}{AED} \\ 
\cline{2-19}
 & 128 & 256 & 128 & 256 & 128 & 256 & 128 & 256 & 128 & 256 & 128 & 256 & 128 & 256 & 128 & 256 & 128 & 256 \\ 
\Xhline{0.6pt}
SimSwap \cite{chen2020simswap} & 24.60\% & 27.09\% & 40.08\% & 31.07\% & 20.02\% & 11.95\% & 45.32\% & 47.99\% & 16.53\% & 15.11\% & 5.58\% & 5.96\% & 13.15\% & 5.47\% & 0.02\% & 0.97\% & 1.31 & 3.55 \\
UniFace \cite{xu2022designing} & 0.48\% & 26.57\% & 11.80\% & 48.27\% & 0.34\% & 31.93\% & 9.17\% & 49.41\% & 6.28\% & 29.85\% & 0.01\% & 2.20\% & 7.24\% & 2.31\% & 0.01\% & 2.44\% & 1.28 & 4.01 \\
CSCS \cite{huang2024identity} & 10.10\% & 10.33\% & 0.29\% & 0.79\% & 0.68\% & 2.35\% & 0.99\% & 1.61\% & 2.30\% & 0.63\% & 0.13\% & 0.56\% & 1.87\% & 0.62\% & 0.00\% & 0.01\% & 1.21 & 3.05 \\
StarGAN-v2 \cite{choi2020stargan} & 5.76\% & 19.97\% & 54.46\% & 39.38\% & 2.35\% & 1.26\% & 15.33\% & 7.91\% & 9.17\% & 3.11\% & 6.13\% & 3.89\% & 5.16\% & 3.51\% & 16.42\% & 8.47\% & 2.56 & 5.51 \\
InfoSwap \cite{gao2021information} & 15.78\% & 18.46\% & 26.92\% & 37.21\% & 6.79\% & 9.21\% & 13.18\% & 5.88\% & 3.35\% & 2.33\% & 5.76\% & 2.31\% & 4.92\% & 3.47\% & 0.22\% & 0.84\% & 1.51 & 3.91 \\
Average & 11.34\% & 20.48\% & 26.71\% & 31.34\% & 6.04\% & 11.34\% & 16.80\% & 22.56\% & 7.53\% & 10.21\% & \underline{3.52\%} & \underline{2.98\%} & 6.47\% & 3.08\% & \textbf{3.33}\% & \textbf{2.55}\% & \textbf{1.58} & \textbf{4.01} \\ 
\Xhline{1.5pt}
\end{tabular}%
}
\end{table*}

\subsection{Visual Quality}
Objectively, the results in Tab.~\ref{tab:imperceptibility} show that our framework achieves high visual quality, surpassing the best-performing baselines such as LampMark \cite{wang2024lampmark} and DiffMark \cite{SUN2026103801}. This confirms our proposed framework's ability to effectively resolve the classic trade-off between payload capacity and fidelity. Crucially, this superior imperceptibility is achieved while embedding a high-capacity 152-D LIDMark, substantially larger than the 30-D or 128-D payloads utilized in prior work. Subjectively, a comparison between $I_{co}$ and $I_{wm}$ in Fig.~\ref{fig:visualization} confirms the absence of discernible artifacts.

\begin{figure}[htbp]
    \centering
    \includegraphics[width=0.46\textwidth]{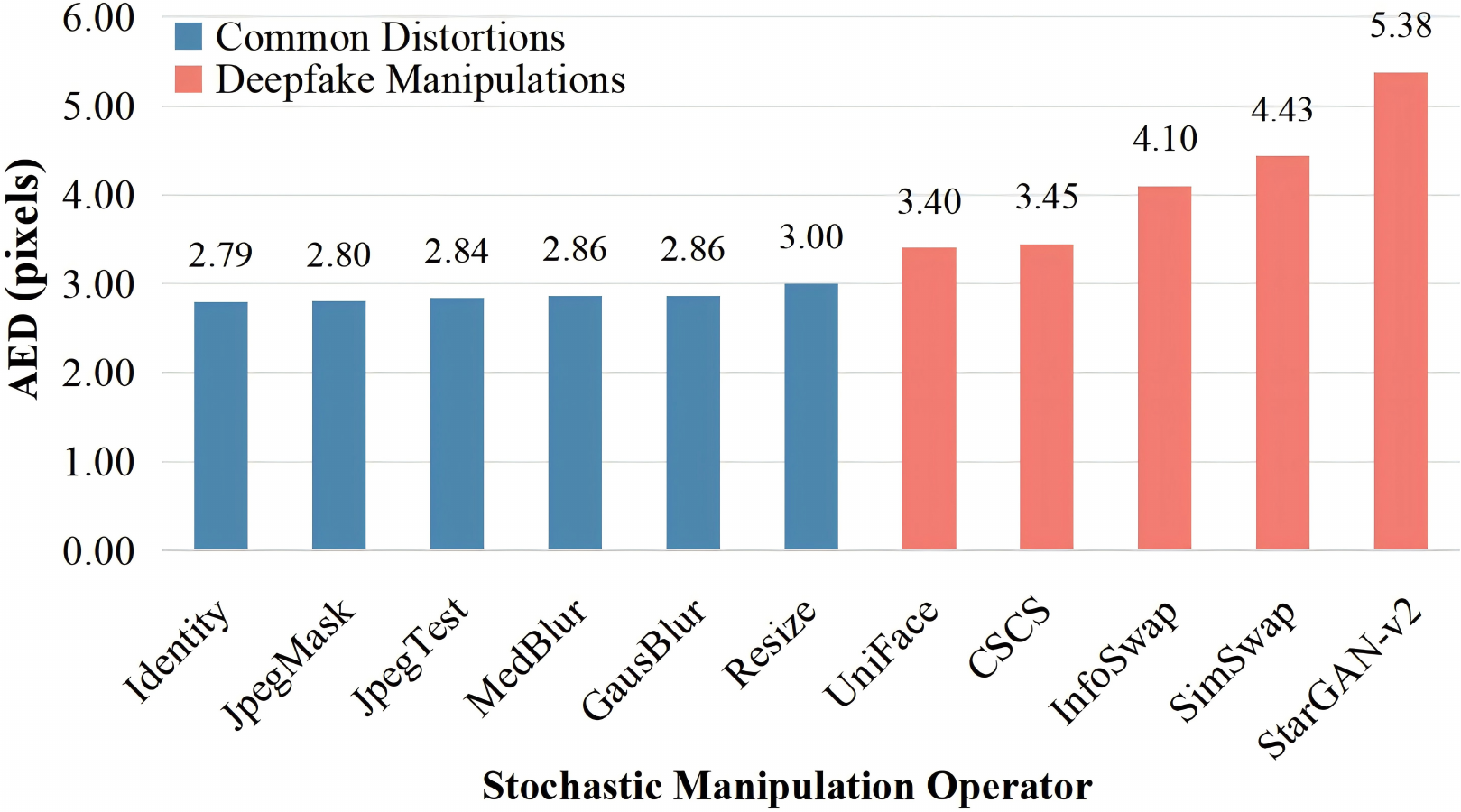}
    \caption{Landmarks Average Euclidean Distance (AED) under various attacks. AED measures the geometric error between FHD-recovered landmarks $\hat{W}_L$ and re-detected landmarks $W_{new}$ on $\mathcal{M}(I_{wm})$. The distribution shows clear separability between benign operations and malicious attacks. The blue $\mathcal{M}_{c}$ bars yield low AED, while the red $\mathcal{M}_{d}$ bars introduce large geometric change, enabling the ``intrinsic-extrinsic'' consistency check.}
    \label{fig:differentiate}
\end{figure}

\subsection{Trifunctional Evaluation}
We evaluate the robustness of the unified 152-D LIDMark by analyzing the recovery of its two components: the 136-D landmarks $\hat{W}_L$ for detection and localization, and the 16-D identifier $\hat{W}_{ID}$ for tracing. The Average Euclidean Distance (AED) measures the recovery quality of the landmarks $\hat{W}_L$. It is calculated in pixel space between predicted points $\hat{p}_{b,i}$ and ground-truth $p_{b,i}$, equivalent to the loss $\mathcal{L}_{L}$ defined in Eq.~\eqref{eq:loss_landmark} scaled by the image size. The Bit Error Rate (BER) measures the recovery quality of the identifier $\hat{W}_{ID}$ by binarizing the predicted logits $\hat{W}_{ID}$ via a sign function $\mathcal{B}(\cdot)$:
\begin{equation}
    \text{BER} = \frac{1}{B \cdot L_{ID}} \sum_{b=1}^{B} \frac{1}{2} \| \mathcal{B}(\hat{W}_{ID, b}) - W_{ID, b} \|_{1}
    \label{eq:metric_ber}
\end{equation}
\textbf{Deepfake Detection and Tampering Localization.} The LIDMark framework achieves its detection and localization capabilities via the ``intrinsic-extrinsic'' consistency check in a fully blind-reference setting, meaning our method requires neither the original image nor ground-truth landmarks during inference. This mechanism validates the geometric integrity of a face by comparing two landmark sets shown in rows 4-6 of Fig.~\ref{fig:visualization}: (1) the intrinsic landmarks $\hat{W}_L$ marked by green dots, robustly recovered by the FHD's regression head, and (2) the extrinsic landmarks $W_{new}$ marked by red dots, re-detected from the manipulated image $\mathcal{M}(I_{wm})$ using the face-alignment library \cite{bulat2017far}. As validated in Tab.~\ref{tab:common_robustness} and Tab.~\ref{tab:deepfake_robustness}, for the geometric signal, the AED remains minimal and stable under common distortions and deepfake attacks. This proves that the FHD's regression head can reliably recover the embedded ``intrinsic'' geometry $\hat{W}_L$ even from the manipulated image. This robustly recovered $\hat{W}_L$ serves as the anchor for our consistency check. For common distortions, the facial geometry remains stable, resulting in a low $\text{AED}(\hat{W}_L, W_{new})$. For deepfake manipulations, the geometry is significantly altered, leading to a high $\text{AED}(\hat{W}_L, W_{new})$. We verified this hypothesis on the validation sets. The AED distributions plotted in Fig.~\ref{fig:differentiate} reveal a separable margin between common distortions and deepfake manipulations. This separability is quantified by an ROC analysis, resulting in a high AUC score of 0.9388, confirming the mechanism's discriminative power. This analysis yields an optimal decision threshold of 3.2375 px, derived via Youden's J statistic. For detection, we apply the check globally: an image is classified as fake if $\text{AED}(\hat{W}_L, W_{new})$ exceeds 3.2375 px. For localization on an image flagged as fake, we apply the check locally, enabled by the semantic ordering of the $W_L$ payload detailed in Fig.~\ref{fig:architecture}(e), which groups landmarks by facial components. Computing the AED for each semantic region against the 3.2375 px threshold pinpoints the manipulated regions.

\begin{table*}[t]
\centering
\caption{Quantitative comparison of BER and AED on LFW \cite{huang2008labeled} under deepfake manipulations $\mathcal{M}_{d}$. All results are reported at $128 \times 128$ and $256 \times 256$ resolutions.}
\label{tab:cross_deepfake_robustness}
\resizebox{\textwidth}{!}{%
\begin{tabular}{l|cc|cc|cc|cc|cc|cc|cc|cccc} 
\Xhline{1.5pt}
\multirow{3}{*}{Manipulation} & \multicolumn{2}{c|}{MBRS \cite{jia2021mbrs}} & \multicolumn{2}{c|}{CIN \cite{ma2022towards}} & \multicolumn{2}{c|}{SepMark \cite{wu2023sepmark}} & \multicolumn{2}{c|}{EditGuard \cite{zhang2024editguard}} & \multicolumn{2}{c|}{LampMark \cite{wang2024lampmark}} & \multicolumn{2}{c|}{DiffMark \cite{SUN2026103801}} & \multicolumn{2}{c|}{KAD-NET \cite{he2025kad}} & \multicolumn{4}{c}{\textbf{Ours}} \\ 
\cline{2-19}
 & \multicolumn{2}{c|}{BER} & \multicolumn{2}{c|}{BER} & \multicolumn{2}{c|}{BER} & \multicolumn{2}{c|}{BER} & \multicolumn{2}{c|}{BER} & \multicolumn{2}{c|}{BER} & \multicolumn{2}{c|}{BER} & \multicolumn{2}{c|}{BER} & \multicolumn{2}{c}{AED} \\ 
\cline{2-19}
 & 128 & 256 & 128 & 256 & 128 & 256 & 128 & 256 & 128 & 256 & 128 & 256 & 128 & 256 & 128 & 256 & 128 & 256 \\ 
\Xhline{0.6pt}
SimSwap \cite{chen2020simswap} & 26.38\% & 27.09\% & 43.02\% & 27.47\% & 25.91\% & 18.07\% & 45.53\% & 47.40\% & 24.95\% & 15.11\% & 9.03\% & 8.35\% & 14.35\% & 6.23\% & 0.65\% & 0.99\% & 1.23 & 3.68 \\
UniFace \cite{xu2022designing} & 0.14\% & 21.16\% & 12.26\% & 47.99\% & 0.41\% & 25.52\% & 8.98\% & 48.80\% & 10.17\% & 29.85\% & 0.01\% & 2.07\% & 7.11\% & 4.36\% & 0.59\% & 2.65\% & 1.88 & 4.22 \\
CSCS \cite{huang2024identity} & 5.27\% & 6.53\% & 0.31\% & 0.70\% & 1.73\% & 1.06\% & 4.29\% & 2.33\% & 10.32\% & 0.63\% & 0.64\% & 3.82\% & 2.20\% & 2.65\% & 0.01\% & 0.01\% & 1.46 & 3.32 \\
StarGAN-v2 \cite{choi2020stargan} & 6.23\% & 19.39\% & 57.98\% & 41.26\% & 3.51\% & 3.34\% & 13.15\% & 3.42\% & 11.10\% & 3.11\% & 7.04\% & 4.11\% & 8.38\% & 4.32\% & 18.12\% & 8.96\% & 2.79 & 5.98 \\
InfoSwap \cite{gao2021information} & 19.77\% & 17.33\% & 28.96\% & 38.62\% & 5.32\% & 13.12\% & 14.22\% & 13.88\% & 3.95\% & 9.64\% & 6.59\% & 4.17\% & 6.12\% & 3.39\% & 0.94\% & 1.21\% & 1.63 & 3.91 \\
Average & 11.56\% & 18.30\% & 28.51\% & 31.21\% & 7.38\% & 12.22\% & 17.23\% & 23.17\% & 12.10\% & 11.67\% & \underline{4.66\%} & \underline{4.50\%} & 7.63\% & 5.24\% & \textbf{4.06}\% & \textbf{2.76}\% & \textbf{1.80} & \textbf{4.22} \\ 
\Xhline{1.5pt}
\end{tabular}%
}
\end{table*}

\begin{table}[t]
\centering
\caption{Objective imperceptibility comparison on LFW \cite{huang2008labeled}. Results are shown across $128 \times 128$ and $256 \times 256$ resolutions.}
\label{tab:cross_imperceptibility}
\setlength{\tabcolsep}{4pt}
\setlength{\heavyrulewidth}{1.5pt}
\begin{tabular}{lcccc} 
\toprule
Model & Image Size & W.L. & PSNR $\uparrow$ & SSIM $\uparrow$ \\
\midrule
MBRS \cite{jia2021mbrs} & $128\times128$ & 30 & 34.99 & 0.90 \\
CIN \cite{ma2022towards} & $128\times128$ & 30 & 39.12 & 0.93 \\
SepMark \cite{wu2023sepmark} & $128\times128$ & 30 & 37.30 & 0.95 \\
EditGuard \cite{zhang2024editguard} & $128\times128$ & 30 & 34.98 & 0.94 \\
LampMark \cite{wang2024lampmark} & $128\times128$ & 30 & 37.18 & 0.94 \\
DiffMark \cite{SUN2026103801} & $128\times128$ & 30 & 39.10 & \underline{0.97} \\
KAD-NET \cite{he2025kad} & $128\times128$ & 30 & \underline{39.21} & 0.96 \\
\textbf{Ours} & $128\times128$ & \textbf{152} & \textbf{39.56} & \textbf{0.97} \\
\midrule
MBRS \cite{jia2021mbrs} & $256\times256$ & 128 & 36.73 & 0.88 \\
CIN \cite{ma2022towards} & $256\times256$ & 128 & 37.21 & 0.82 \\
SepMark \cite{wu2023sepmark} & $256\times256$ & 128 & 38.28 & 0.93 \\
EditGuard \cite{zhang2024editguard} & $256\times256$ & 128 & 34.23 & 0.84 \\
LampMark \cite{wang2024lampmark} & $256\times256$ & 128 & 39.56 & 0.95 \\
DiffMark \cite{SUN2026103801} & $256\times256$ & 128 & \underline{40.33} & \underline{0.97} \\
KAD-NET \cite{he2025kad} & $256\times256$ & 128 & 39.98 & 0.94 \\
\textbf{Ours} & $256\times256$ & \textbf{152} & \textbf{40.68} & \textbf{0.98} \\
\bottomrule
\end{tabular}
\end{table}

\textbf{Source Tracing.} The FHD's classification head reconstructs the 16-D identifier $\hat{W}_{ID}$ for immediate provenance verification. As validated in Tab.~\ref{tab:common_robustness}, our framework achieves the lowest average BER across baselines under common distortions. Furthermore, Tab.~\ref{tab:deepfake_robustness} shows that many baselines such as CIN \cite{ma2022towards} and EditGuard \cite{zhang2024editguard} fail catastrophically against severe deepfake manipulations with BERs exceeding 30-50\%. Facing such attacks, our framework again achieves the best overall average BER, maintaining near-perfect recovery against identity-focused swaps like CSCS \cite{huang2024identity}. Unlike identity swaps, attribute-based modifications such as StarGAN-v2 \cite{choi2020stargan} alter the image globally, extending beyond the facial region to modify background and structural textures. This global alteration introduces severe signal-level distortions that significantly complicate watermark recovery, an inherent vulnerability also seen in DiffMark \cite{SUN2026103801}. Nevertheless, our average performance across diverse threat models remains consistently effective.

\subsection{Cross-Dataset Evaluation}
We test generalizability by applying the model trained on CelebA-HQ \cite{karras2017progressive} directly to the unseen LFW dataset \cite{huang2008labeled}, which features different identities and capture conditions. Tab.~\ref{tab:cross_imperceptibility} shows that high fidelity is maintained. Furthermore, the intrinsic landmark AED remains low, proving the framework's strong robustness to out-of-distribution data. Crucially, Tab.~\ref{tab:cross_deepfake_robustness} confirms its tracing robustness, achieving the lowest average BER among all baselines on deepfakes.

\begin{table}[t]
\centering
\caption{Ablation study on the FHD loss components. Symbols $\checkmark$ and \ding{55} denote loss enablement and disablement, respectively. The results for the $128 \times 128$ and $256 \times 256$ resolutions are shown in the top and bottom sections of the table.}
\label{tab:ablation_hyperparams}
\setlength{\tabcolsep}{2pt}
\begin{tabular}{cc|cc|cc|cc}
\Xhline{1.5pt}
\multirow{2}{*}{$\lambda_{L}$} & \multirow{2}{*}{$\lambda_{ID}$} & \multirow{2}{*}{PSNR $\uparrow$} & \multirow{2}{*}{SSIM $\uparrow$} & \multicolumn{2}{c|}{BER $\downarrow$} & \multicolumn{2}{c}{AED $\downarrow$} \\
\cline{5-6} \cline{7-8}
& & & & $\mathcal{M}_{c}$ & $\mathcal{M}_{d}$ & $\mathcal{M}_{c}$ & $\mathcal{M}_{d}$ \\
\Xhline{0.8pt}
% --- 128x128 Block ---
\ding{55} & $\checkmark$ & 42.12 & 0.98 & 0.00\% & 0.14\% & 342.39 & 359.98 \\
$\checkmark$ & \ding{55} & 47.31 & 0.99 & 0.42\% & 0.49\% & 2.31 & 2.79 \\
$\checkmark$ & $\checkmark$ & 40.22 & 0.98 & 0.27\% & 3.33\% & 1.17 & 1.58\\
\Xhline{0.8pt}
% --- 256x256 Block ---
\ding{55} & $\checkmark$ & 43.54 & 0.99 & 0.00\% & 0.29\% & 540.91 & 405.97 \\
$\checkmark$ & \ding{55} & 49.18 & 0.99 & 0.35\% & 0.48\% & 5.29& 5.65 \\
$\checkmark$ & $\checkmark$ & 44.31 & 0.99 & 0.10\% & 2.55\% & 3.06 & 4.01 \\
\Xhline{1.5pt}
\end{tabular}
\end{table}

\subsection{Ablation Studies}
\label{sec:ablation_studies}
We ablate decoder loss components, $\mathcal{L}_L$ and $\mathcal{L}_{ID}$, to validate the FHD design. Results in Tab.~\ref{tab:ablation_hyperparams} reveal their strong interdependence. Disabling landmark loss causes geometric recovery to completely fail, confirming that $\mathcal{L}_L$ is essential for detection and localization. Conversely, disabling ID loss severely compromises tracing. The full jointly optimized model best balances all objectives. It gains landmark recovery capability while accepting a minor trade-off in deepfake tracing BER, proving that the multi-task loss and FHD are both necessary for the trifunctional forensic goal.
\section{Conclusion}
We propose the LIDMark framework, the first proactive forensics approach to unify deepfake detection, tampering localization, and source tracing. This ``all-in-one'' solution is enabled by the high-capacity 152-D LIDMark and a novel FHD. From a single backbone, the FHD extracts landmarks for detection and localization via an ``intrinsic-extrinsic'' consistency check, alongside identifiers for tracing. Extensive experiments demonstrate that our framework provides a robust, imperceptible and unified paradigm for facial forensics. Future work will enhance robustness against unseen deepfakes and extend to video modalities.

{
    \small
    \bibliographystyle{ieeenat_fullname}
    \bibliography{main}
}

\setcounter{section}{0} 
\renewcommand{\thesection}{\Alph{section}}
\clearpage
\setcounter{page}{1}
\maketitlesupplementary

\section{Scalability Analysis}
\label{sec:Scalability_Analysis}
\begin{figure*}[t]
    \centering
    \includegraphics[width=\linewidth]{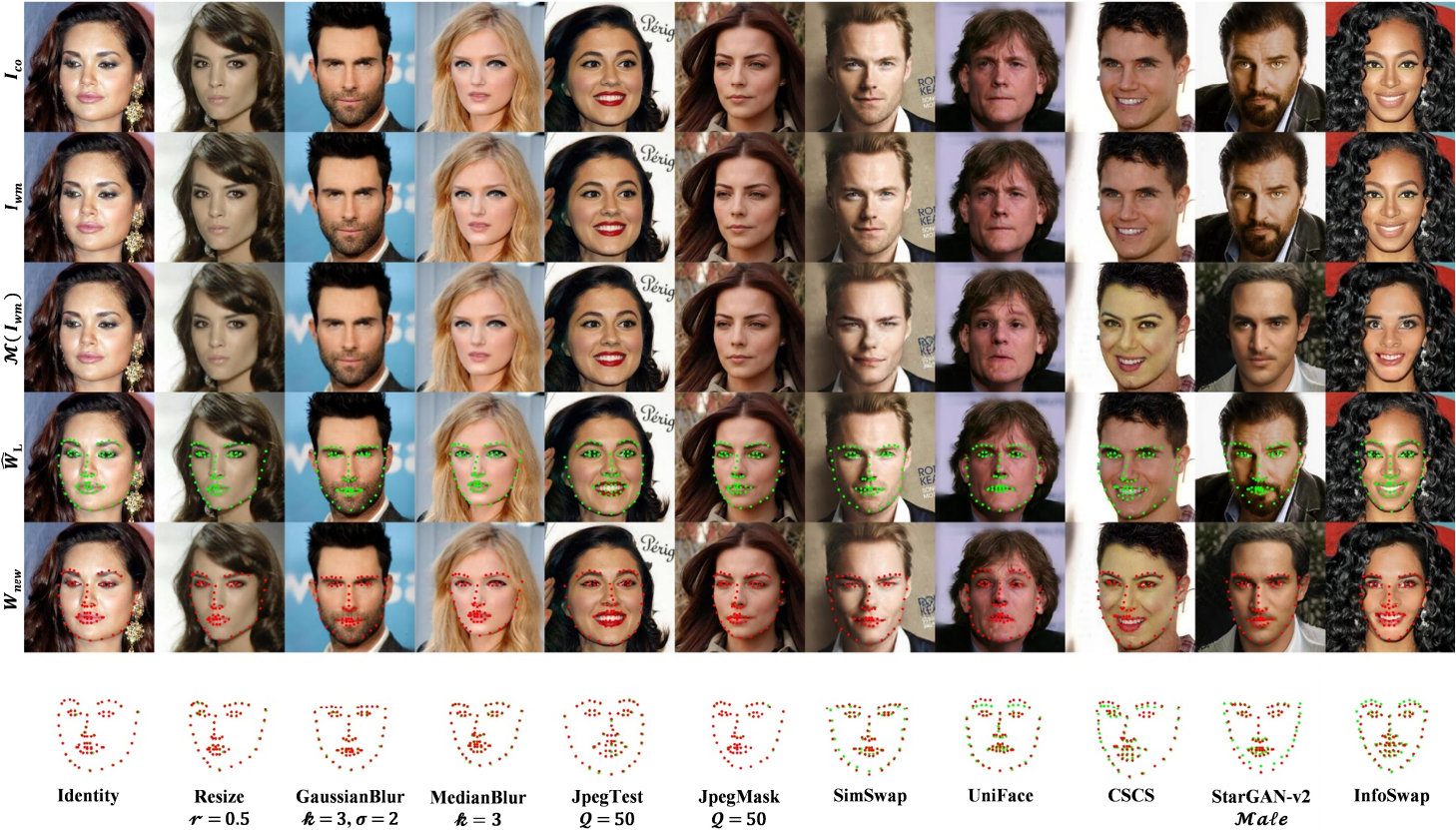}
    \caption{Visual assessment of the 168-D LIDMark robustness and imperceptibility. Comparing rows 1 and 2 shows the watermarked image $I_{wm}$ is indistinguishable from the cover image $I_{co}$. Row 3 displays the manipulation results $\mathcal{M}(I_{wm})$. The ``intrinsic-extrinsic'' consistency check compares the green dots in row 4, representing FHD-recovered intrinsic landmarks $\hat{W}_L$, against the red dots in row 5, representing re-detected extrinsic landmarks $W_{new}$. Row 6 merges these two landmark sets into a combined image to visualize their spatial differences.
    }
    \label{fig:168-visualization}
\end{figure*}

Since the 136-D landmark vector $W_L$ is already sufficient for detection and localization, we assess LIDMark's scalability by expanding the identifier payload $W_{ID}$ from 16-D to 32-D. While a 16-D payload supports $2^{16}$ unique identities and is sufficient for most scenarios, a 32-D payload accommodates $2^{32} \approx 4.3 \times 10^9$ identities to cater to large-scale applications. This expansion increases the total payload to 168-D, comprising the 136-D landmark and the new $32$-D identifier. While $W_{ID}$ can represent arbitrary provenance or copyright data, our experiments, for convenience, use the hashed filename to simulate this tracing scenario. Results are detailed in Tabs.~\ref{tab:imperceptibility_supp}-\ref{tab:168_deepfake_merged} and Fig.~\ref{fig:168-visualization}. The framework accommodates a $10.5\%$ increase in total payload capacity with only a marginal PSNR cost, while yielding a more stable and robust recovery of the geometric landmark signal against both common distortions and advanced deepfake manipulations. Notably, the expanded identifier creates no interference with the geometric signal, confirming that our factorized decoding strategy effectively disentangles these two heterogeneous tasks. This validates the potential to expand $W_{ID}$ for more complex or granular source tracing.

\begin{table}[ht]
\centering
\caption{Objective imperceptibility comparison on CelebA-HQ \cite{karras2017progressive}. Results are at $256 \times 256$ resolution. We compare the standard payload (W.L. 152) against the expanded (W.L. 168).}
\label{tab:imperceptibility_supp} 
\begin{tabular}{lccc} 
\Xhline{1.5pt}
W.L. & Image Size & PSNR $\uparrow$ & SSIM $\uparrow$ \\
\Xhline{0.6pt}
\textbf{152} & $256 \times 256$ & 44.31 & 0.99 \\
\textbf{168} & $256 \times 256$ & 42.36 & 0.99 \\
\Xhline{1.5pt}
\end{tabular}
\end{table}

\begin{table}[ht]
\centering 
\caption{Quantitative comparison on CelebA-HQ \cite{karras2017progressive} of BER and AED for LIDMark with different capacities (W.L. 152 vs. 168) under common distortions $\mathcal{M}_{c}$, demonstrating highly robust scalability.}
\label{tab:168_common_merged} 
\resizebox{\columnwidth}{!}{%
\begin{tabular}{l|cc|cc} 
\Xhline{1.5pt}
\multirow{2}{*}{Distortion} & \multicolumn{2}{c|}{\textbf{W.L. 152}} & \multicolumn{2}{c}{\textbf{W.L. 168}} \\
\cline{2-5}
 & BER $\downarrow$ & AED $\downarrow$ & BER $\downarrow$ & AED $\downarrow$ \\
\Xhline{0.6pt}
Identity & 0.00\% & 2.96 & 0.00\% & 2.14 \\
Resize & 0.00\% & 3.00 & 0.00\% & 2.17 \\
GausBlur & 0.00\% & 3.03 & 0.00\% & 2.16 \\
MedBlur & 0.00\% & 3.01 & 0.02\% & 2.16 \\
JpegTest & 0.57\% & 3.39 & 0.00\% & 2.36 \\
JpegMask & 0.01\% & 2.98 & 0.01\% & 2.17 \\
Average & \textbf{0.10}\% & \textbf{3.06} & \textbf{0.00}\% & \textbf{2.19} \\
\Xhline{1.5pt}
\end{tabular}
}
\end{table}

\begin{table}[t]
\centering 
\caption{Quantitative comparison on CelebA-HQ \cite{karras2017progressive} of BER and AED for LIDMark with different capacities (W.L. 152 vs. 168) under deepfake manipulations $\mathcal{M}_{d}$, demonstrating highly robust scalability.} 
\label{tab:168_deepfake_merged} 
\resizebox{\columnwidth}{!}{%
\begin{tabular}{l|cc|cc} 
\Xhline{1.5pt}
\multirow{2}{*}{Manipulation} & \multicolumn{2}{c|}{\textbf{W.L. 152}} & \multicolumn{2}{c}{\textbf{W.L. 168}} \\
\cline{2-5}
 & BER $\downarrow$ & AED $\downarrow$ & BER $\downarrow$ & AED $\downarrow$ \\
\Xhline{0.6pt}
SimSwap \cite{chen2020simswap} & 0.97\% & 3.55 & 0.59\% & 2.58 \\
UniFace \cite{xu2022designing} & 2.44\% & 4.01 & 0.18\% & 2.47 \\
CSCS \cite{huang2024identity} & 0.01\% & 3.05 & 0.00\% & 2.21 \\
StarGAN-v2 \cite{choi2020stargan} & 8.47\% & 5.51 & 14.33\% & 5.32 \\
InfoSwap \cite{gao2021information} & 0.84\% & 3.91 & 1.19\% & 2.48 \\
Average & \textbf{2.55}\% & \textbf{4.01} & \textbf{3.26}\% & \textbf{3.01} \\
\Xhline{1.5pt}
\end{tabular}%
}
\end{table}

\section{Complexity Analysis}
Tab.~\ref{tab:cost} presents a comprehensive benchmarking of model complexity, contrasting LIDMark with existing dual-function methods. In terms of parameter efficiency, LIDMark demonstrates superior compactness. This is largely attributed to our unified design, where a single shared backbone efficiently extracts rich and versatile features for multiple downstream tasks, avoiding the structural redundancy of dual-branch encoders. At a resolution of $128 \times 128$, our model requires only 5.82M parameters, representing approximately 20\% and 17\% of the capacity of SepMark \cite{wu2023sepmark} and WaveGuard \cite{11224900}, respectively. This lightweight footprint minimizes storage requirements, making LIDMark exceptionally suitable for deployment on storage-constrained edge devices and mobile platforms.
\begin{table}[t]
\centering
\caption{Comparison of model complexity against dual-function baselines. We report the number of Parameters (M) and FLOPs (G). For fair comparison, the watermark length (W.L.) is standardized to 152-D for all methods.}
\label{tab:cost}
\setlength{\heavyrulewidth}{1.5pt}

\resizebox{\linewidth}{!}{
    \begin{tabular}{lccc} 
    \toprule
    Model & Image Size & Params (M) & FLOPs (G) \\
    \midrule
    SepMark \cite{wu2023sepmark} & $128\times128$ & 28.310 & 4.031 \\
    WaveGuard \cite{11224900} & $128\times128$ & 34.810 & 27.014 \\
    KAD-NET \cite{he2025kad} & $128\times128$ & 8.542 & 7.835 \\
    \textbf{Ours} & $128\times128$ & 5.815 & 12.221 \\
    \midrule
    SepMark \cite{wu2023sepmark} & $256\times256$ & 28.310 & 16.059 \\
    WaveGuard \cite{11224900} & $256\times256$ & 34.810 & 27.011 \\
    KAD-NET \cite{he2025kad} & $256\times256$ & 48.542 & 31.357 \\
    \textbf{Ours} & $256\times256$ & 21.023 & 60.622 \\
    \bottomrule
    \end{tabular}
}
\end{table}

Regarding computational cost, we observe that LIDMark incurs higher FLOPs. This increased computational density is a deliberate and necessary investment to achieve trifunctional capability. Unlike baselines that are limited to coarse-grained detection or tracing, our FHD performs dense, pixel-level feature processing to ensure precise tampering localization and high-fidelity watermark recovery. While methods like SepMark achieve lower FLOPs, they sacrifice the granularity required for localization and the robustness needed for challenging deepfake attacks. Consequently, we argue that the marginal increase in inference latency is a worthy exchange for the significant gains in forensic security and the unification of three distinct tasks within a single framework.

\end{document}